\documentclass[10pt,twocolumn,letterpaper]{article}

\usepackage[pagenumbers]{wacv} 

\usepackage{graphicx}
\usepackage{amsmath}
\usepackage{amssymb}
\usepackage{booktabs}
\usepackage{enumitem}
\usepackage{multicol}
\usepackage{multirow}

\usepackage{tikz}
\usepackage{pgfplots}
\usepackage{verbatim}
\usepackage{cuted}
\usepackage{subcaption}

\usepackage{pifont}
\newcommand{\cmark}{\ding{51}}%
\usepackage[pagebackref,breaklinks,colorlinks]{hyperref}

\usepackage[capitalize]{cleveref}
\crefname{section}{Sec.}{Secs.}
\Crefname{section}{Section}{Sections}
\Crefname{table}{Table}{Tables}
\crefname{table}{Tab.}{Tabs.}

\def\wacvPaperID{2426} 
\def\confName{WACV}
\def\confYear{2024}

\begin{document}

\title{Stereo Matching in Time: 100+ FPS Video Stereo Matching for Extended Reality}

\author{Ziang Cheng$^{1,2*}$, \;\;Jiayu Yang$^{1,2*}$,\;\; Hongdong Li$^{2}$\\
$^1$Tencent XR Vision Labs, $^2$Australian National University \\
{\tt\small \{ziang.cheng, jiayu.yang, hongdong.li\}@anu.edu.au}
}

\maketitle

\begin{strip}
  \centering
  \vspace{-1.5cm}
\includegraphics[width=1.0\linewidth]{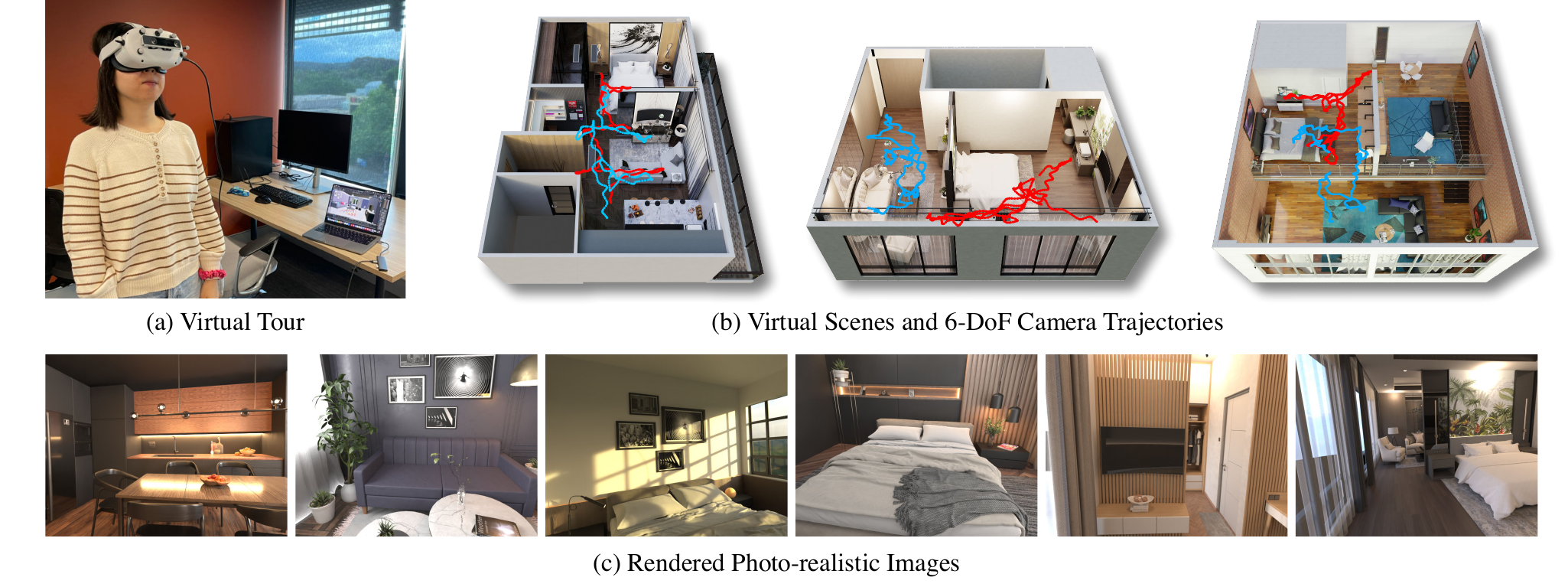}\\
  \vspace{-0.3cm}
  \captionof{figure}{\noindent\textbf{The XR-Stereo dataset:} (a) We collected high-fidelity camera movement trajectories by taking virtual tours of virtual scenes and recording 6-DoF camera poses from VR/AR HMDs. (b) Examples of the captured virtual scenes and 6-DoF camera trajectories. (c) Examples of rendered photo-realistic images.} 
  \vspace{-0.3cm}
  \label{fig:dataset1}
\end{strip}

\begin{abstract}
   Real-time Stereo Matching is a cornerstone algorithm for many Extended Reality (XR) applications, such as indoor 3D understanding, video pass-through, and mixed-reality games. Despite significant advancements in deep stereo methods, achieving real-time depth inference with high accuracy on a low-power device remains a major challenge. One of the major difficulties is the lack of high-quality indoor video stereo training datasets captured by head-mounted VR/AR glasses.  To address this issue, we introduce a novel video stereo synthetic dataset that comprises photorealistic renderings of various indoor scenes and realistic camera motion captured by a 6-DoF moving VR/AR head-mounted display (HMD). This facilitates the evaluation of existing approaches and promotes further research on indoor augmented reality scenarios. Our newly proposed dataset enables us to develop a novel framework for continuous video-rate stereo matching.
      
   As another contribution, our dataset enables us to proposed a new video-based stereo matching approach tailored for XR applications, which achieves real-time inference at an impressive 134fps on a standard desktop computer, or 30fps on a battery-powered HMD. Our key insight is that disparity and contextual information are highly correlated and redundant between consecutive stereo frames. By unrolling an iterative cost aggregation in time (i.e. in the temporal dimension), we are able to distribute and reuse the aggregated features over time. This approach leads to a substantial reduction in computation without sacrificing accuracy. We conducted extensive evaluations and comparisons and demonstrated that our method achieves superior performance compared to the current state-of-the-art, making it a strong contender for real-time stereo matching in VR/AR applications. 

\end{abstract}

\section{Introduction}

Extended reality (or XR in short) is a collective term that refers to immersive technologies, including Virtual Reality (VR), Augmented Reality (AR) and Mixed Reality (MR).  The applications for XR are vastly growing, ranging from gaming and entertainment to education, healthcare, and business. \let\thefootnote\relax\footnote{$^*$Equal contribution; Work done during an internship at Tencent.} \let\thefootnote\relax\footnote{$^\dagger$ Work completed when with Tencent.}

Real-time Stereo Matching is key algorithm running on a VR/AR headset, which enables a wide range of applications such as visual passthrough and 3D mapping.  Efficient stereo matching is particularly important for stand-alone (untethered) VR/AR headsets, as these are typically low-power mobile devices with limited computing power. Therefore, any computational overhead from stereo matching algorithms can reduce the responsiveness of human interaction and quickly drain the headset's battery, ultimately diminishing the overall user experience. Hence, it is imperative to develop highly efficient stereo algorithms that can provide accurate estimations in real-time on such low-power devices. One promising direction is to utilize temporal information. For high-frequency video stereo matching, consecutive frames can have large overlapped region. The redundant overlapped region in the temporal dimension can be utilized to reduce the computation overhead per frame.

A significant roadblock to the development of aforementioned algorithms is the absence of high-quality indoor stereo video datasets tailored for XR scenarios. Our examination of the current stereo datasets reveals several limitations that inhibit our exploration of temporal redundancy for efficient video stereo matching in XR scenarios.

\noindent\textbf{Insufficient indoor environments.} Indoor scenarios present unique challenging effects, including large texture-less area such as wall or floor, transparent windows and mirror reflections. Standard stereo datasets such as SceneFlow \cite{mayer2016large} were not built for indoor stereo matching.

\noindent\textbf{Absence of video sequences.} Among the few dataset that provide indoor scenarios, most are targeted on single-frame stereo matching or lack accurate camera poses (\eg \cite{wang2019irs,bao2020instereo2k}), which considerably diminishes the capabilities of stereo methods, hindering the exploration of temporal relations between consecutive frames. 

\noindent\textbf{Lacking photo-realism.} For the few possible options, we find~\cite{tartanair2020iros} suffers subpar texture and shading quality, while~\cite{schops2017multi} only provide gray-scale images. The diversity of indoor environments is also limited in both datasets, making it difficult to generalize to the wider indoor XR domain.

In this paper, we propose two novel contributions to facilitate the development of real-time stereo matching and other 3D vision tasks in XR scenarios. Our first contribution is a novel indoor XR video stereo dataset, which comprises photo-realistic stereo video sequences of various indoor environments rendered by a physically-based path tracing rendering engine, along with realistic and complex 6-DoF camera trajectories captured by VR/AR head-mounted displays (HMDs). Furthermore, the dataset includes various complete and accurate ground-truth labels, such as disparity, depth, optical flow, normal \etc. Our novel video stereo dataset enables us to propose our second contribution, which is a video stereo matching algorithm that can achieve 100+fps inference speed on a standard desktop computer while maintaining comparable precision to current state-of-the-art methods. We achieve this by exploring the computational redundancy in the temporal dimension, based on the key insight that disparity and contextual information are highly correlated between consecutive stereo frames. Specifically, we assume known camera pose, and unroll an iterative cost aggregation network into the temporal dimension through temporal warping, which enables us to distribute and reuse the aggregated feature over time and leads to a substantial reduction in computation without sacrificing accuracy. Extensive experiments demonstrate that our method achieves significant improvements in inference speed compared to current state-of-the-art methods while maintaining comparable precision, making it a strong contender for real-time stereo matching in XR applications.

\noindent\textbf{Our contributions} are summarized as follows:
\begin{itemize}[noitemsep]
\item A high-fidelity synthetic dataset for video stereo matching in indoor XR scenarios, consisting of photo-realistic stereo sequences and 6-DoF realistic camera motion captured by VR/AR head-mounted displays.
\item A video stereo matching pipeline that exploits temporal redundancy of scene geometry to achieve real-time inference with high accuracy.
\item Extensive experiments and comparisons with existing methods show that our method can maintain accuracy comparable to state-of-the-art methods while achieving an impressive 134fps inference speed on a desktop computer.
\end{itemize}

\section{Related work}

\noindent\textbf{Video stereo datasets.}
A huge roadblock for developing efficient video stereo algorithms for indoor XR scenario is the lack of high-quality video stereo datasets in indoor environment. Specifically, we find several limitations of existing stereo datasets, including (1) Deficiency of indoor scenarios, (2) absence of video sequence and (3) lacking photo-realism, that largely blocked our way to explore the utilization of temporal information redundancy to develop a highly efficient video stereo matching algorithms dedicated for indoor XR scenario (see Tab. \ref{tab:datasets} for a summary). Existing stereo datasets are collected from various domains, including autonomous driving~\cite{Geiger2012CVPR,huang2019apolloscape}, robotics~\cite{schops2017multi,tartanair2020iros}, movie~\cite{Butler:ECCV:2012} or synthesized from 3D models for general stereo matching training~\cite{mayer2016large}. Indoor scenarios present unique challenging effects, including large texture-less ares such as wall or floor, transparent windows and mirror reflections. Few existing datasets~\cite{wang2019irs,schops2017multi,tartanair2020iros,scharstein2002taxonomy,scharstein2014high} provide indoor scenes for training stereo matching to overcome these challenges. Among them, most are targeted on single-frame stereo matching~\cite{scharstein2002taxonomy,scharstein2014high} and do not provide high-quality real-time video sequences, which considerably diminishes the capabilities of current stereo methods, limiting them to single-frame setups and hindering the exploration of temporal relations. For the very few stereo datasets that offer indoor stereo video sequences, \cite{wang2019irs} do not provide camera pose. Only~\cite{schops2017multi,tartanair2020iros} provide accurate camera poses. Lack of camera data can ultimately limit the exploration of utilizing temporal relation, as camera pose is required for accurately model the relation between pixels in consecutive video frames. For these two possible options, we find~\cite{tartanair2020iros} exhibits subpar rendering quality and lacks the photo-realistic effects present in indoor XR environments, while~\cite{schops2017multi} only provide low resolution gray-scale images. Their variety of indoor environments is also limited for training algorithms dedicated to the indoor XR domain.

To facilitate the development of real-time stereo matching and other 3D vision tasks in XR scenarios, we propose a novel indoor video stereo dataset that utilizes physically-based path tracing to achieve photorealistic rendering of stereo pairs, along with \emph{real} 6-DoF movement captured by a head-mounted display (HMD) as the camera trajectory. Our dataset is designed and implemented to a high standard to facilitate the evaluation of our approach and promote further research in Extended Reality scenarios.

\noindent\textbf{Stereo Matching.}
Stereo Matching is a long-standing task in computer vision, involving the estimation of the disparity between a stereo image pair. Classical methods~\cite{scharstein2002taxonomy,hirschmuller2005accurate,bleyer2011patchmatch} compute hand-crafted matching costs along with local, semi-global, or global cost aggregation to achieve accurate, complete, and smooth disparity estimation. In recent years, deep learning-based stereo methods~\cite{xu2021bilateral,zhang2019ga,guo2019group,tosi2021smd,mao2021uasnet} have achieved superior results by utilizing learned image features to build cost volumes, learned cost aggregation networks, and learned disparity regression. 

A recent approach, RAFT-Stereo\cite{lipson2021raft}, employs a GRU structure to iteratively lookup matching costs and refine disparity estimation. Another approach, CREStereo~\cite{li2022practical}, extends this method with a cascade framework and adaptive correlation. 

\begin{table*}[t]
\vspace{-0.5cm}
\centering
\resizebox{\linewidth}{!}{
\begin{tabular}{l|c|c|c|c|c|c|c}\hline
\textbf{Dataset}      & \textbf{Year} & \textbf{Type}      & \textbf{Scenario}       & \textbf{Camera Pose} & \textbf{Camera Motion} & \textbf{Video}   & \textbf{Disparity}        \\\hline\hline
Sintel\cite{Butler:ECCV:2012}       & 2012 & Synthetic & Movie        & Virtual         & Virtual Camera           & 50 frames     & Rendered         \\\hline
KITTI Stereo\cite{Geiger2012CVPR} & 2012 & Real      & Road           & Estimated   & Driving       & N/A     & LiDAR            \\\hline
KITTI VO\cite{Geiger2012CVPR}     & 2012 & Real      & Road           & Estimated   & Driving       & 10 Hz   & LiDAR            \\\hline
Middlebury\cite{scharstein2014high}   & 2014 & Real      & Laboratory     & N/A         & N/A           & N/A     & Structured Light \\\hline
SceneFlow\cite{mayer2016large}    & 2016 & Synthetic & Various        & Virtual     & Spline        & N/A     & Rendered         \\\hline
ETH3D\cite{schops2017multi}        & 2017 & Real      & Indoor/Outdoor & Estimated   & Robotics      & 13.6 Hz & Laser Scan       \\\hline
Apollo\cite{huang2019apolloscape}       & 2018 & Real      & Road           & Estimated   & Driving       & 30 Hz   & LiDAR            \\\hline
TartanAir\cite{tartanair2020iros}    & 2020 & Synthetic & Various        & Virtual     & Robotics      & 10Hz-30Hz     & Rendered         \\\hline
IRS\cite{wang2019irs}          & 2021 & Synthetic & Indoor         & Virtual     & Spline        & N/A     & Rendered         \\\hline
XR-Stereo (Ours)         & 2023 & Synthetic & \textbf{Indoor XR}      & \textbf{HMD Recorded}    & \textbf{Human Head Movement}     & \textbf{30 Hz}   & Rendered        \\
\hline
\end{tabular}
}
\vspace{-0.3cm}
\caption{Stereo Matching Datasets. Our proposed XR-Stereo dataset focus on indoor extended reality (XR) scenario and provide 6-DoF camera motion recorded by VR/AR HMD through our virtual tour pipeline. }
\label{tab:datasets}
\vspace{-0.5cm}
\end{table*} 

\noindent\textbf{Real-time Stereo Matching.}
Few existing stereo matching methods focus on achieving real-time inference with high accuracy. Classical methods, such as PatchMatch Stereo~\cite{bleyer2011patchmatch}, are capable of running in real-time, but their estimation accuracy is far inferior to current learning-based methods due to their hand-crafted cost metric and cost aggregation. These methods are largely outperformed by deep learning-based counterparts. Few recent deep learning-based methods have explored real-time stereo matching~\cite{duggal2019deeppruner,lipson2021raft}. DeepPruner~\cite{duggal2019deeppruner} is based on PatchMatch Stereo~\cite{barnes2010generalized,bleyer2011patchmatch}, which implements a differentiable PatchMatch layer for pruning the disparity searching space in low computational cost to alleviate the computational overhead of subsequent deep cost volume and deep cost aggregation network. StereoNet~\cite{khamis2018stereonet} propose to use hierarchical refinement to improve efficiency. HiTNet~\cite{tankovich2021hitnet} use a multi-resolution initialization paired with coarse-to-fine slanted window based propagation to improve efficiency. Coex~\cite{bangunharcana2021correlate} propose a Guided Cost Volume Excitation to alleviate the computation of 3D convolutions. Most closely related to our method is RAFT-Stereo~\cite{lipson2021raft}, which utilizes a multilevel recurrent field to iteratively refine disparity estimation. It iteratively looks up matching cost from a pre-built cost volume and uses an iterative cost aggregation network to gradually refine the disparity estimation. Such a strategy has shown great promise in terms of accuracy and generality, but its runtime increases linearly with the increasing number of iterations. Therefore, it has to trade-off accuracy by reducing the number of iterations to improve inference speed.

\noindent\textbf{Video Stereo.}
Temporal information can substantially contribute to both the accuracy and efficiency of stereo matching; however, recent deep stereo methods rarely explore the utilization of temporal information. Classical methods like Patchmatch Stereo~\cite{bleyer2011patchmatch} utilize temporal relation by propagating previous disparity estimations to the current frame as disparity hypotheses but lack further utilization of contextual information over time. Open-World Stereo~\cite{zhong2018open} builds an LSTM connecting latent features along the time dimension. However, they heavily focus on the unsupervised open-world stereo setup and provide very limited insight or evaluation on the utilization of temporal information for improving stereo matching accuracy or efficiency. DeepVideoMVS~\cite{duzceker2021deepvideomvs} adopted similar LSTM structure for temporal multi-view stereo, where the temporal matching require a static scene that is infeasible in XR scenario. Very recently, a concurrent work TemporalStereo~\cite{zhang2022temporalstereo} also explored the utilization of temporal information in stereo matching. They extend a single-frame coarse-to-fine estimation framework similar to cascade stereo methods \cite{gu2020cascade,yang2020cost}, and warp previous cost volume into the current view with hand-crafted statistical fusion. They also warp previous disparity to the current time frame for extra disparity hypothesis. Based on these simple temporal extensions, their temporal model yields incremental accuracy improvement over their single-frame baseline but is slower at runtime. Unlike TemporalStereo, our framework is specifically designed to reduce the computational cost per frame by learning temporal cost aggregation. Our model can better utilize temporal information and can achieve 3x to 5x faster inference speed with superior accuracy than our single-frame baseline.

\section{XR-Stereo Dataset}\label{sec:dataset}
We first introduce our new video stereo synthetic dataset, XR-Stereo, which contains 60K stereo images in 640x480 resolution. We design our dataset in an indoor extended reality (XR) setup, where the stereo cameras are mounted on a head mounted display (HMD). Below we introduce details of this new dataset. 

\subsection{Indoor Scenes} 
We use a set of 13 carefully crafted, photo-realistic 3D virtual indoor environments. As shown in Figure \ref{fig:dataset1}(b), the virtual scenes selected for this research primarily focus on household environments such as living rooms, kitchens, bedrooms, and study rooms, while also featuring a few additional environments such as office, hospital and hotel room. 

\subsection{Recorded 6-DoF Camera Trajectory} 
We aim to obtain accurate camera poses of realistic movement trajectories by capturing the motion of a virtual reality/augmented reality (VR/AR) head-mounted display (HMD) as its wearers walk through virtual scenes. To accomplish this, we have developed a virtual tour pipeline that enables users to explore virtual scenes while wearing a VR/AR HMD. As illustrated in Figure \ref{fig:dataset1}(a), the pipeline consists of a real-time stereoscopic viewport rendering engine that is connected to the VR/AR HMD using OpenXR APIs. The pipeline utilizes the 6-DoF head pose of the HMD in real-time and associates it with the virtual head location in the virtual scene. The corresponding stereoscopic viewport images of the virtual head location are then rendered in real-time and streamed back to the HMD. This allows the wearer to freely explore the virtual scene while physically walking in the real world, subject to the physical space available for walking. Figure \ref{fig:dataset1}(b) shows examples of captured head trajectories overlaid on virtual scene examples. To ensure a diverse set of trajectories, we collected trajectories from different users with varying movement styles for each virtual scene, for a total of 17 trajectories.

\subsection{Physically-based rendered images} Existing synthetic datasets typically utilize real-time rendering engine such as Lumen \cite{lumen} or Eevee \cite{blender} for its high efficiency and low computational cost. However, they are sub-optimal for training XR-oriented stereo matching networks due to the lack of challenging real-world photometric effects (such as specular reflections and transmitted lights) that are present in indoor environments. To simulate complex real-world optics, we use Blender's path tracing engine Cycles~\cite{blender}. All the virtual scenes are shaded with Physics-Based Rendering materials for photometric fidelity. With this setup, we are able to synthesize a diverse range of optical effects in high fidelity, including mirror reflection, refraction, subsurface scattering and secondary reflection. Physically-based path tracing consumes significantly more computational resources over its real-time counterparts. We implement our rendering pipeline using 80 distributed computing nodes, each node containing 4 NVIDIA Tesla V100 GPUs. Rendering the entire dataset took around two weeks.

\subsection{Lens and lighting effects}
Our dataset draws inspiration from the SceneFlow dataset~\cite{MIFDB16} and provide rendering of two sets of images, which we refer to as cleanpass and finalpass. The cleanpass set serves as a baseline, providing a clear and unadulterated view of the scene radiance, while the finalpass set contains real-world lens and lighting effects such as including motion-blur, defocus, rolling-shutter, lens glare and indoor light flickering. 

\subsection{Data Type and Potential Applications} We provide various types of data and ground-truth, including pixel-wise raw light intensity, RGB image, rendered disparity, rendered depth, optical flow, surface normal, visual ray vectors, surface mesh, instance labels, object bounding box, diffuse/specular/transmission layer separation etc. These data can facilitate the development of various indoor XR applications, such as indoor 3D reconstruction or Visual See-through (VST), and generic computer vision tasks such as monocular depth estimation, stereo matching, multi-view stereo, 3D reconstruction, structure from motion, visual odometry, SLAM, etc.

\subsection{Limitation} 
The current version of our XR-Stereo dataset does not contain moving objects. In XR scenarios, various applications require modeling of moving objects in the scene, such as pets or people. Our dataset also lacks of an egocentric virtual human model, particularly the hand and body. The human model will aid in the development of hand tracking, human motion estimation, digital twin, and other related applications. 

\begin{figure}[t]
  \centering
  \vspace{-0.3cm}
\includegraphics[width=0.9\linewidth]{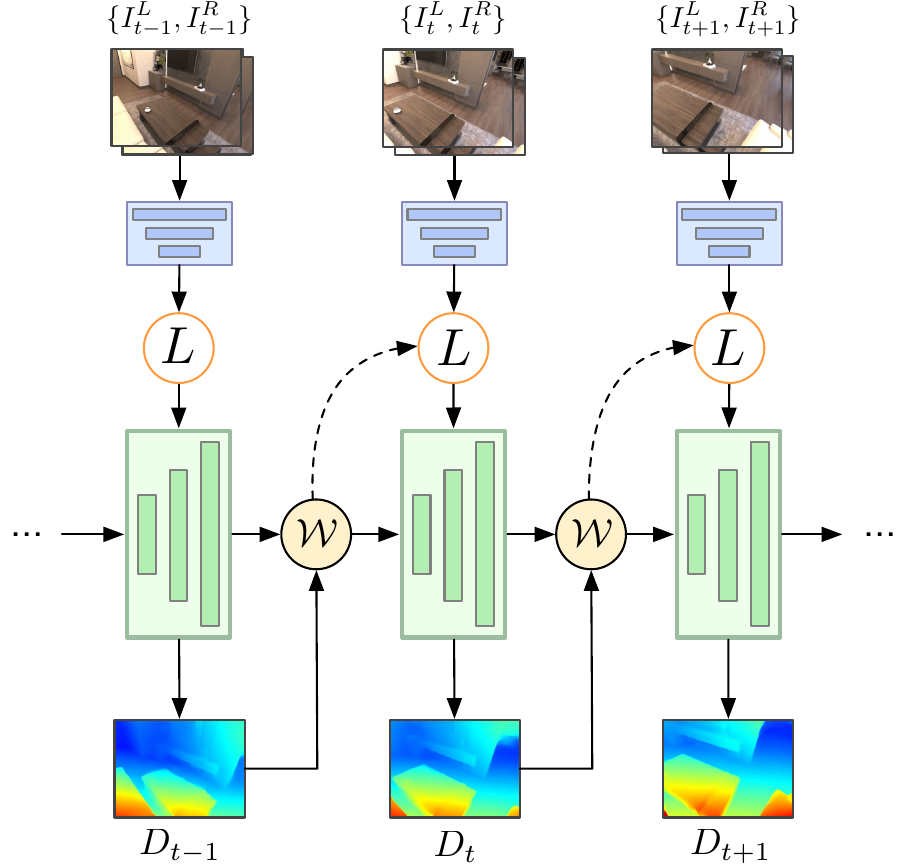}\\
  \vspace{-0.1cm}
  \caption{Real-time video stereo matching framework. We unroll an iterative cost aggregation network into temporal dimension, which enables us to distribute and reuse the aggregated feature over time and leads to substantial reduction in computation without sacrificing accuracy. In this example, we perform a single GRU iteration per frame.}
  \label{fig:pipeline}
  \vspace{-0.5cm}
\end{figure}

\section{Continuous video stereo matching}

We now introduce our video stereo matching framework. Our key idea is to leverage the temporal redundancy of video within a RAFT-style cost aggregation scheme, where disparities are iteratively looked-up and refined by a GRU. Instead of running multiple iterations per frame in a frame-by-frame manner, we warp the previous frame disparity hypothesis to current frame to warm-start the GRU. As illustrated in Fig \ref{fig:pipeline}, this strategy significantly reduces GRU iterations per frame.

For time step $t$ with new stereo image frames inputs $\{I_t^L,I_t^R\}$, their relevant world-to-camera pose inputs $\{P_t^L,P_t^R\}$, their camera intrinsic parameters $\{K^L,K^R\}$ and stereo baseline $B$, our method estimate disparity $D_t$ for the left stereo image. We first extract matching feature and contextual feature from input images~(Sec.~\ref{sec:feature_extraction}). We then warp previous disparity estimation and previous hidden state into current camera frame~(Sec.~\ref{sec:temporal_warping}). Based on the warped disparity, we perform disparity lookup and compute matching cost on-the-fly. The matching cost is used along with current contextual feature in a recurrent network to estimate disparity for current time frame~(Sec.~\ref{sec:look_up})). We train this network in supervised manner~(Sec.~\ref{sec:sup}). 

\subsection{Feature Extraction}\label{sec:feature_extraction}
For any time step $t$ with stereo image inputs $\{I_t^L,I_t^R\}$, we firstly extract matching feature $\{F_t^L,F_t^R\}$ and context feature $\{C_t^L,C_t^R\}$ using a shared-weight feature extraction network $\phi_f: I\rightarrow \{F,C\}$. We adopt the feature extraction network from \cite{lipson2021raft} for fair comparison. 

\subsection{State Initialization}\label{sec:init}
If the given stereo inputs on time step $t$ is the very first frame, we perform a disparity and hidden state initialization. We use a small network $\phi_{0}$ to estimate an initial disparity taking current left context feature as input. We initialize hidden state as all zeros. Pose of previous frame is set to be identical to current frame. 

\begin{figure}[t]
  \centering
  \vspace{-0.2cm}
\includegraphics[width=0.7\linewidth]{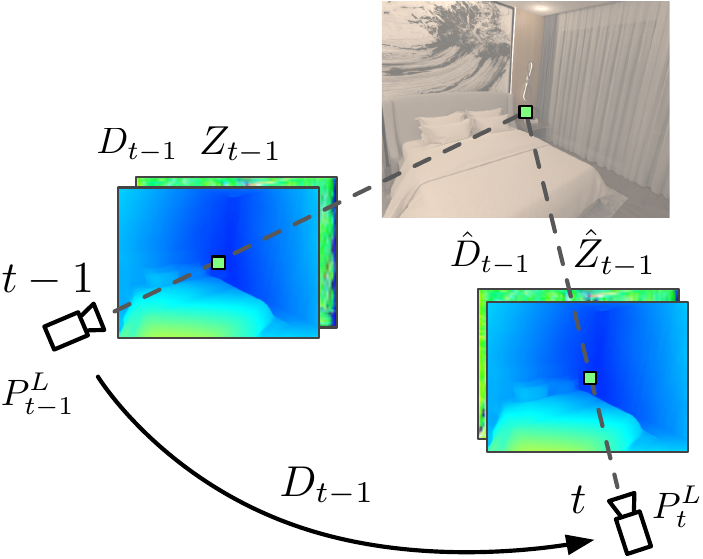}\\
  \vspace{-0.1cm}
  \caption{Temporal Warping. We reuse previous disparity estimation and hidden state feature by warping them to current frame based on relative camera pose.}
  \vspace{-0.3cm}
  \label{fig:warp}
\end{figure}

\subsection{Temporal Warping}\label{sec:temporal_warping}
We warp previous disparity estimation $D_{t-1}$ and GRU hidden state $Z_{t-1}$ into current camera frame, termed as $\hat D_{t-1}, \hat Z_{t-1}$ accordingly. As illustrated in Fig \ref{fig:warp}, this warping address the view point change caused by camera movement based on multi-view geometry. Specifically, we compute a transformation matrix $T^{geo}_{t-1,t}$ that maps stereo pixels $(u,v,d)$ from previous left camera coordinate to current left camera coordinate. 

\begin{equation}
T^{geo}_{t-1,t} = Q[RT]_{t-1,t}Q^{-1},
\end{equation}

where $[RT]_{t-1,t}$ is the relative camera pose and $Q$ is the stereo transformation from stereo coordinates $(u,v,d)$ to camera coordinates $(x,y,z)$.

\begin{equation}
Q = \begin{bmatrix}
1 & 0 & 0 & -c_x \\
0 & 1 & 0 & -c_y \\
0 & 0 & 0 & f \\
0 & 0 & 1/b & 0 \\
\end{bmatrix}
\end{equation}

We warp previous disparity and hidden state to current frame using this transformation $T^{geo}_{t-1,t}$, resulting in $\hat D^{geo}_{t-1}, \hat Z^{geo}_{t-1}$ respectively. To handle occlusion during forward warping, we use Softmax Splatting~\cite{niklaus2020softmax} with weight proportional to current disparity, so that when multiple locations from previous view are mapped to the same location at current view, the nearest one takes priority. In case of disocclusion (holes), all missing values are assigned zero. We implement forward warping as a non-parametric layer in our network that is differentiable \wrt hidden state $Z_{t-1}$.

\subsection{Disparity look-up and cost aggregation}\label{sec:look_up}
Upon each new input frame at timestamp $t$, the cost aggregation GRU takes as initial input the warped hidden state of previous frame $\hat Z_{t-1}$ and warped disparity $\hat D_{t-1}$, and performs iterative cost look-ups and disparity updates conditioned on context features. Same as RAFT-Stereo \cite{lipson2021raft}, we use feature correlation as cost metric to compute photometric matching cost $M_t \in \mathbb{R}^{h\times w\times K}$ as

\begin{equation}
M_t(u,v,d) = F^L_t(u,v)\cdot F^{R}_t(u+d,v).
\end{equation}

Contrast to RAFT-Stereo~\cite{lipson2021raft}, we are able to significantly reduce the number of GRU iterations. Even with a single GRU iteration per frame, our method performs comparably to RAFT-Stereo~\cite{lipson2021raft} using 20 GRU iterations. 

\begin{table*}[!t]
\vspace{-0.3cm}
\begin{center}

\resizebox{0.6\linewidth}{!}{
\begin{tabular}{ll|cccc|c}
\hline
\multicolumn{2}{c|}{Method} & EPE & D1 & D3 & D5 & FPS\\
\hline\hline
\parbox[t]{2mm}{\multirow{7}{*}{\rotatebox[origin=c]{90}{Single-frame}}}
& HiTNet*\cite{tankovich2021hitnet} & 4.51 & 36.4 & 24.1 & 11.34 & 19.48 \\
& GA-Net-deep\cite{Zhang2019GANet} & 2.11 & 22.56 & 11.05 & 8.46 & 1.24 \\
& GWC-Net\cite{guo2019group} & 1.87 & 20.45 & 10.01 & 7.99 & 8.10 \\
& PCW-Net\cite{shen2022pcw} & 1.77 & 19.43 & 9.96 & 7.84 & 6.53 \\
& ACV-Net\cite{xu2022attention} & 1.69 & 19.31 & 8.94 & 6.57 & 13.89 \\
& RAFT-Stereo\cite{lipson2021raft}~\small{(RT @ 7 iters)} & 1.93 & 20.93 & 9.85 & 7.03 & 49.83 \\
& RAFT-Stereo\cite{lipson2021raft}~\small{(RT @ 20 iters)} & 1.70 & 18.18 & 8.37 & 6.03 & 19.97\\
\hline
\parbox[t]{2mm}{\multirow{5}{*}{\rotatebox[origin=c]{90}{Video}}}
& DeepVideoMVS*  & 2.20 & 20.2 & 9.52 & 7.10 & 13.22 \\
& Ours-Fast & 1.67 & 19.40 & 8.86 & 6.25 & \textbf{134.05} \\
& Ours~\small{(1 iter)} & 1.48 & 18.64 & 8.55 & 5.95 & 108.09 \\
& Ours~\small{(2 iters)} & 1.44 & 16.80 & 7.89 & 5.50 & 90.97\\
& Ours~\small{(5 iters)} & \textbf{1.42} & \textbf{16.31} & \textbf{7.69} & \textbf{5.36} & 57.46 \\
\hline
\end{tabular}
}
\end{center}
\vspace{-0.4cm}
\caption{\textbf{XR-Stereo dataset.} Our method outperforms all existing methods in terms of both accuracy and runtime fps. For accuracy similar to RAFT-Stereo~\cite{lipson2021raft} real-time model \small{(RT @ 20 iters)}, Our fast model achieved an impressive 134 fps on inference. }
\label{tab:main}
\vspace{-0.1cm}
\end{table*}

\subsection{Supervised Training}\label{sec:sup}
We train our network in a supervised manner. Following common practice in stereo matching methods \cite{guo2019group,lipson2021raft}, we use the $l_1$ distance between estimated disparity and final disparity as disparity loss $\mathcal{L}_d$.

\begin{equation}
    \mathcal{L}_d = ||D-D_{gt}||_1
\end{equation}

Unlike RAFT-Stereo \cite{lipson2021raft} that applies an exponential loss weight over iteration steps, we treat the disparity output of each time step equally important and do not weigh down outputs on early time steps.

\section{Experiments}
In this section, we demonstrate the performance of our approach with a comprehensive set of experiments. Below, we describe the datasets and benchmarks, the implementation details, presentation and analysis of our results.

\subsection{Datasets}
We conduct extensive experiments on our proposed XR-Stereo dataset and also verify the performance of our method on the real-world KITTI VO dataset \cite{Geiger2012CVPR}. 

\noindent\textbf{XR-Stereo dataset} is our newly proposed synthetic indoor video stereo dataset. It consists the rendering of 14 virtual indoor scenes, which forms more than 60k photo-realistic stereo image pairs of indoor scenario. We use $640\times 480$ resolution and 30Hz video frame rate. We split the dataset into 2 validation scenes and 2 testing scenes, and the rest are used for training.

\begin{figure}[t]
\begin{center}
\resizebox{0.8\linewidth}{!}{
\begin{tikzpicture}
    \begin{axis}[
        xlabel=Inference Time (ms),
        ylabel=EPE,
        xmin=0, xmax=60,
        ymin=1.25, ymax=3.6,
        xtick={0, 10,20,30,40,50,60},
        ytick={1.5,2,2.5,3,3.5},
        width=\linewidth,
        height=0.7\linewidth
        ]
        
    \addplot[mark=square*,teal] plot coordinates {
        (7.46 ,1.67) 
    };
    \addlegendentry{Ours-Fast}
    
    \addplot[smooth,mark=*,blue] plot coordinates {
        (9.25 ,1.4796112458654145) 
        (10.99 ,1.4355372168729355) 
        (13.35 ,1.4283921078185124) 
        (15.44 ,1.4089602729814155) 
        (17.403 ,1.4204825803984593) 
        
    };
    \addlegendentry{Ours}

    \addplot[smooth,color=red,mark=x]
        plot coordinates {
            (8.08 ,3.3764276603826193) 
            (10.25 ,2.7406532059728654) 
            (12.2 ,2.4381193938189765) 
            (16.12 ,2.097394011375209) 
            (20.08 ,1.9258693374175124) 
            (26.31 ,1.8073008830848425) 
            (36.9 ,1.7286114285745326) 
            (50.1 ,1.703528229458276) 
            
        };
    \addlegendentry{RAFT-Stereo RT}
    \end{axis}
    \end{tikzpicture}
}
\end{center}
\vspace{-0.3cm}
\caption{\noindent\textbf{XR-Stereo dataset.} Inference time and accuracy comparison between our method and our single-frame baseline RAFT-Stereo\cite{lipson2021raft} real-time model. Our model outperform RAFT-Stereo real-time model on both accuracy and inference speed.}
\label{fig:runtime_epe}
\vspace{-0.2cm}
\end{figure}
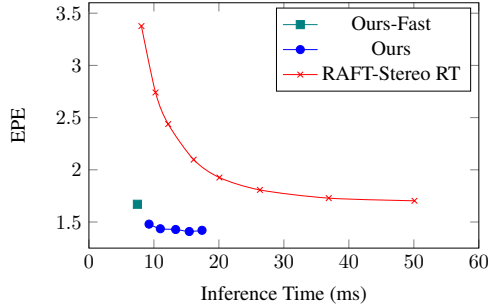

\noindent\textbf{KITTI Visual Odometry (VO) dataset}~\cite{Geiger2012CVPR} consists of image sequences acquired from a car driving in urban scenarios, captured with a calibrated stereo camera system and a high-precision GPS/IMU localization system. It cover a wide range of scenarios, such as residential areas, highways, and urban centers. It provides synchronized RGB stereo images and LiDAR measurements, IMU measurements, and GPS localization data. The dataset is widely used for evaluating visual odometry, stereo matching, and SLAM algorithms. We train our model from scratch and split the dataset into 20 training videos and 2 test videos.

\begin{figure*}[t]
  \centering
\includegraphics[width=0.85\linewidth]{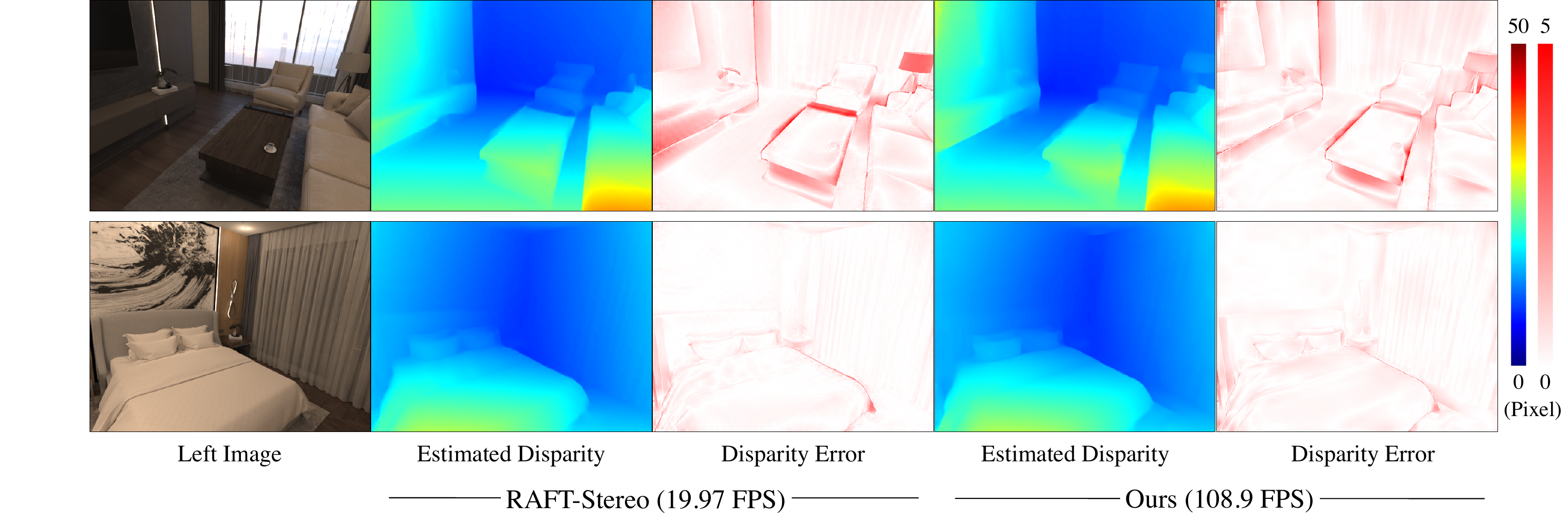}\\
  \vspace{-0.2cm}
  \caption{\noindent\textbf{XR-Stereo dataset.} Qualitative results}
  \vspace{-0.3cm}
  \label{fig:results}
\end{figure*}

\subsection{Metrics}
We use standard evaluation metrics to assess our results. These metrics include the average end-point error (EPE), as well as the percentage of pixels with disparity error greater than 1 pixel (D1), 3 pixels (D3), and 5 pixels (D5). Furthermore, we also compute the bad 1\%, bad 3\%, and bad 5\% EPE on the KITTI VO dataset, which correspond to the top 1\%, 3\%, and 5\% percentiles of EPE, respectively.

\subsection{Implementation Details}
We implement two variants of our method, a full model (Ours) and a fast model (Ours-Fast). For the full model, we include results from our model with 1, 2 and 5 GRU iterations per frame. The fast model is introduced to maximize inference speed, for which we run GRU once per frame and disable the temporal warping. The fast model differs from our full model in two ways: (a) we remove the temporal warping by assuming camera motion between consecutive frames is small and continuous (b) we half intermediate feature channels within the feature encoder. We train both full model and fast model using Adam optimizer for 250K iterations using batch size 8 and sequence length 16. We use eight NVIDIA V100 32G GPUs for training. For all evaluations, we use a PC with one NVIDIA RTX 3090 TI GPU. To deploy our trained model on HMD, we convert the fast model via ONNX to a device-friendly float-point format running on Qualcomm XR2 chip without quantization.

\subsection{Comparison with Existing Methods}

We first compare our method with related methods on our XR-Stereo dataset. All methods are trained on our dataset unless otherwise stated, and are grouped into single-frame (Single-frame, or S in short) or video stereo methods (Video). For single frame methods we mainly compare to our real-time baseline RAFT-Stereo~\cite{lipson2021raft}. For video stereo methods we only compare to DeepVideoMVS~\cite{duzceker2021deepvideomvs} as other methods~\cite{zhong2018open,zhang2022temporalstereo} did not release code. Results are listed in Tab. \ref{tab:main}. 
HiTNet~\cite{tankovich2021hitnet} did not release training code, so we can only use its released pre-trained model for evaluation. Disparity from DeepVideoMVS~\cite{duzceker2021deepvideomvs} were obtained by $d = \frac{b\times f}{Z}$. DeepVideoMVS runs much slower than reported in original paper due to our dataset has 4X higher image resolution than ScanNet \cite{dai2017scannet}.
Our fast model, which lacks the temporal warping, already outperforms all existing methods in EPE and achieved an impressive 134 FPS inference speed. Our full model that runs a single GRU iteration maintains a high inference speed at 100+ FPS. By further adding GRU iterations, our method consistently outperformed all competing methods on all metrics.

\begin{table*}[!t]
\vspace{-0.5cm}
\begin{center}

\resizebox{0.8\linewidth}{!}{
\begin{tabular}{ll|ccccccc}
\hline
\multicolumn{2}{c|}{Method} & EPE & D1 & D3 & D5 & Bad 1\% & Bad 3\% & Bad 5\% \\
\hline\hline
\parbox[t]{2mm}{\multirow{2}{*}{\rotatebox[origin=c]{90}{S}}}
& RAFT-Stereo\cite{lipson2021raft}~\small{(RT @ 7 iters)} & 1.82 & 26.47 & 10.26 & 6.71 & 26.93 & 13.69 & 8.27 \\
& RAFT-Stereo\cite{lipson2021raft}~\small{(RT @ 20 iters)} & 1.77 & 25.76 & 9.79 & 6.39 & 27.01 & 13.30 & 7.85 \\
\hline
\parbox[t]{2mm}{\multirow{4}{*}{\rotatebox[origin=c]{90}{Video}}}

& Ours-Fast & 1.91 & 30.58 & 11.92 & 7.51 & 25.15 & 13.55 & 8.64 \\
& Ours~\small{(1 iters)} & 1.84 & 30.33 & 11.14 & 6.93 & 24.87 & 12.82 & 7.94 \\
& Ours~\small{(2 iters)} & 1.72 & 26.71 & 10.11 & 6.47 & \textbf{24.71} & 12.59 & 7.55 \\
& Ours~\small{(5 iters)} & \textbf{1.66} & \textbf{24.52} & \textbf{9.57} & \textbf{6.22} & 24.87 & \textbf{12.40} & \textbf{7.32} \\

\hline
\end{tabular}
}
\end{center}
\vspace{-0.4cm}
\caption{\textbf{KITTI VO dataset.} Performance of our method compared with RAFT-Stereo~\cite{lipson2021raft} in real-world scenario. }
\label{tab:kitti_odometry}
\vspace{-0.3cm}
\end{table*}

\subsection{Efficiency versus accuracy}
Apart from the general inference fps provided in Table \ref{fig:results}, we also specifically compared the runtime efficiency and accuracy of our model to the single-frame baseline RAFT-Stereo. For both models, we plot the runtime-EPE curve with varying number of GRU iterations. The results are shown in Figure \ref{fig:runtime_epe}. Our model achieves significantly better accuracy compared to our single-frame counterpart while operating within a fraction of its time. 

\subsection{Generalization Ability}
We evaluate the generalization ability of our model on KITTI Visual Odometry (VO) dataset with low frame rate and sparse depth supervision. Results are shown in Tab. \ref{tab:kitti_odometry}. Our model outperforms the single-frame baseline using significantly less GRU iterations, which proves its generalization ability to real-world applications.

\subsection{Ablation study}
We provide ablation experiments on the XR-Stereo dataset to evaluate the contribution of the proposed modules and also provide detailed analysis of our method.

\noindent\textbf{Temporal warping}
In this ablation we evaluate the importance of correct geometric alignment by temporal warping. A baseline model is trained without warping and compared to our full model running one GRU iteration per frame. Results are show in Tab. \ref{tab:ablation}. The performance of baseline model deteriorates as camera motion increases when the full model is robust to large motions. Similar observations are made in Fig.\ref{fig:movement}, where the fast model is vulnerable to large motions due to lack of temporal warping.

\begin{figure}[t]
\begin{center}
\resizebox{0.8\linewidth}{!}{
\begin{tikzpicture}
    \begin{axis}[
        xlabel=Pose Noise Level,
        ylabel=EPE,
        xmin=0, xmax=10,
        ymin=1.4, ymax=2.6,
        xtick={1,3,5,7,9},
        ytick={1.5,1.75,2,2.25,2.5},
        width=\linewidth,
        height=0.5\linewidth
        ]
    \addplot[smooth,mark=*,blue] plot coordinates {
        (0 ,1.4699720531192035) 
        (1 ,1.5114497191625217) 
        (2 ,1.5176707911847453) 
        (3 ,1.6085645801205772) 
        (4 ,1.69554387048097) 
        (5 ,1.7829168292709583) 
        (6 ,1.883822316258102) 
        (7 ,1.980760042545743) 
        (8 ,2.0893492100734075) 
        (9 ,2.1955451566440862) 
        (10 ,2.3020382201146483) 
    };

    \end{axis}
    \end{tikzpicture}
}
\end{center}
\vspace{-0.6cm}
\caption{\noindent\textbf{XR-Stereo dataset.} EPE versus Pose Noise. Each noise level adds an increment of 0.3 degrees maximum rotation noise and 1mm maximum translation noise.}
\label{fig:noise}
\end{figure}
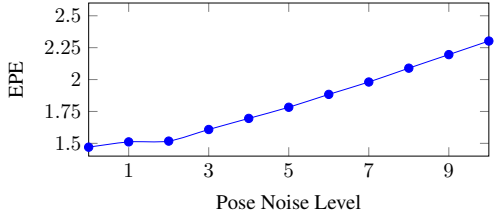

\noindent\textbf{Pose noise}
We analyze the robustness of our model against noisy camera pose. We manually add a random noise into to the input pose before using it in our temporal warping modules. Specifically, we add a random rotation and translation noise drawn from uniform distributions. Results are show in Fig. \ref{fig:noise}, where each noise level adds an increment of 0.3 degrees maximum rotation noise and 1mm maximum translation noise. Our model is affected by extreme pose noise. However, since our method only requires relative pose between two consecutive frames, many SLAM pipelines falls within noise level 1. \Eg a modified ORB-SLAM3\cite{ORBSLAM3_TRO} runs within 0.3 degree and 0.5mm inter-frame noise on our headset, with local bundle adjustment.

\begin{table}[t]
\vspace{-0.2cm}
\scriptsize
\begin{center}
\resizebox{0.9\columnwidth}{!}{
\begin{tabular}{c|cc|cc}
\hline
\multirow{2}{*}{Temporal Warping} & \multicolumn{2}{c|}{1X speed} & \multicolumn{2}{c}{6X speed} \\ 
 & EPE & D1 & EPE & D1
\\ \hline\hline
& 1.70 & 18.29 & 3.18 & 32.05 \\ \hline
\cmark & 1.48 & 18.64 & 1.68 & 21.16 \\ \hline

\end{tabular}
}
\end{center}
\vspace{-0.4cm}
\caption{\textbf{XR-Stereo dataset.} Contribution of temporal warping.}
\label{tab:ablation}
\vspace{-0.5cm}
\end{table}

\noindent\textbf{Movement speed}
Since our model is specifically designed to utilize the relation between consecutive frames, inter-frame motion can affect our model's performance. High speed motions reduce overlap between consecutive frames and can further introduce occlusion/disocclusion. We increase inter-frame motion intensity by skipping frames in a 30Hz input stereo video stream, simulating up to 20x original speed. Results are shown in Fig. \ref{fig:movement}. Our full model is robust to high speed motions when the fast model (without warping) is vulnerable to such changes.

\noindent\textbf{Performance curve on initialization}
We now analyze the performance curve on model initialization. Since our model rely on temporal iterative cost aggregation, in a fresh start, it require certain amount of frames to be processed to reach a stable performance. In Fig. \ref{fig:init}, we show the disparity accuracy curve as more frames are fed into our model (1 iter) upon initialization. The disparity accuracy of the very first few frames gradually improved along with more iterations of temporal cost aggregation been performed. The performance of our model stabilized after around 15 frames, which roughly corresponds to a 0.5 second video duration.

\begin{figure}[t]
\begin{center}
\resizebox{0.8\linewidth}{!}{
\begin{tikzpicture}
    \begin{axis}[
        xlabel=Init. Iterations,
        ylabel=$EPE$,
        xmin=0, xmax=30,
        ymin=0, ymax=1,
        xtick={5,10,15,20,25,30},
        ytick={0.2,0.4,0.6,0.8},
        width=\linewidth,
        height=0.5\linewidth
        ]
    \addplot[smooth,mark=*,blue] plot coordinates {
        (1 ,0.8607890605926514) 
        (2 ,0.5818251967430115) 
        (3 ,0.4948809742927551) 
        (4 ,0.5099976658821106) 
        (5 ,0.46537700295448303) 
        (6 ,0.4587145447731018) 
        (7 ,0.43322640657424927) 
        (8 ,0.40766221284866333) 
        (9 ,0.39049139618873596) 
        (10 ,0.37565430998802185) 
        (11 ,0.3622429668903351) 
        (12 ,0.35551178455352783) 
        (13 ,0.3410736322402954) 
        (14 ,0.3157568573951721) 
        (15 ,0.29770246148109436) 
        (16 ,0.2864713668823242) 
        (17 ,0.2946004271507263) 
        (18 ,0.28764769434928894) 
        (19 ,0.2914232611656189) 
        (20 ,0.2921583354473114) 
        (21 ,0.2877786159515381) 
        (22 ,0.2906060814857483) 
        (23 ,0.28720614314079285) 
        (24 ,0.2788906693458557) 
        (25 ,0.27282392978668213) 
        (26 ,0.26918768882751465) 
        (27 ,0.28311410546302795) 
        (28 ,0.288907527923584) 
        (29 ,0.2912512719631195) 
        (30 ,0.2974795401096344) 
    };

    \end{axis}
    \end{tikzpicture}
}
\end{center}
\vspace{-0.5cm}
\caption{\noindent\textbf{XR-Stereo dataset.} EPE curve as our model progresses through an input video. Error converges after 15 stereo frames.}
\vspace{-0.1cm}
\label{fig:init}
\end{figure}
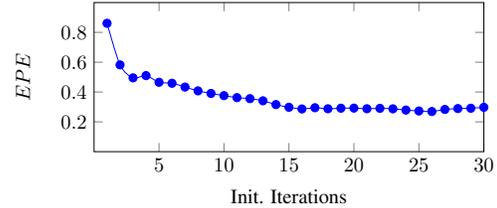
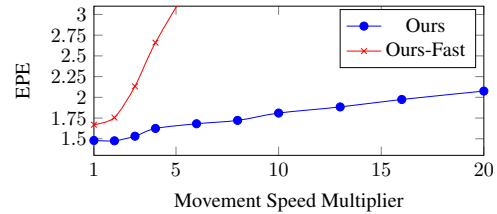
\begin{figure}[t]
\vspace{-0.2cm}
\begin{center}
\resizebox{0.8\linewidth}{!}{
\begin{tikzpicture}
    \begin{axis}[
        xlabel=Movement Speed Multiplier,
        ylabel=EPE,
        xmin=1, xmax=20,
        ymin=1.3, ymax=3.1,
        xtick={1,5,10,15,20},
        ytick={1.5,1.75,2,2.25,2.5,2.75,3},
        width=\linewidth,
        height=0.5\linewidth
        ]
    \addplot[smooth,mark=*,blue] plot coordinates {
        (1 ,1.4796179170026547) 
        (2 ,1.4754625135803705) 
        (3 ,1.531200556070349) 
        (4 ,1.624712591347927) 
        (6 ,1.6813329467151918) 
        (8 ,1.721665863900245) 
        (10 ,1.8102685615090253) 
        (13 ,1.884091784883084) 
        (16 ,1.9741840910806199) 
        (20 ,2.0753493734902766) 
    };
    \addlegendentry{Ours}

    \addplot[smooth,color=red,mark=x]
        plot coordinates {
        (1 ,1.6696672688463075) 
        (2 ,1.755719981492699) 
        (3 ,2.1329774033887303) 
        (4 ,2.6601624785910083) 
        (6 ,3.487304451614477) 
        (8 ,4.172243618040899) 
        (10 ,5.192499024486994) 
        (13 ,5.747945137857411) 
        (16 ,6.212265646517879) 
        (20 ,6.746224036164088) 
        (25 ,7.429588564854908) 
        (30 ,7.391651569102048) 
        (40 ,7.325937645615272) 
    };
    \addlegendentry{Ours-Fast}
    \end{axis}
    \end{tikzpicture}
}
\end{center}
\vspace{-0.5cm}
\caption{\noindent\textbf{XR-Stereo dataset.} Model robustness with respect to movement speed. Our fast model, which lacks temporal warping, can be largely affected by high movement speed. Our full model with temporal warping achieve reasonable performance even in extreme speed (20x).}
\label{fig:movement}
\vspace{-0.5cm}
\end{figure}

\subsection{Limitation}
Our proposed approach, while effective in stereo matching scenarios with stereo video inputs, may face limitations in applications where such inputs are not available. Besides, it may not provide reliable and accurate disparity estimation at the very first stereo frame, thereby limiting its use in on-demand stereo matching applications. 

\section{Conclusion}
We present two novel contributions to facilitate the development of XR, including an indoor XR video stereo dataset implemented in high-fidelity and a highly efficient real-time video stereo matching framework that can potentially run in real-time on low-power stand-alone VR/AR headsets. In the future, we would like to improve our dataset to enable the development of more XR algorithms, and extend our video stereo~framework~to~sceneflow estimation.

{\small
\bibliographystyle{ieee_fullname}
\bibliography{egbib}
}

\end{document}